# Topic-based Evaluation for Conversational Bots

**Fenfei Guo,**[*,1] **Angeliki Metallinou,**[2] **Chandra Khatri,**[2]
**Anirudh Raju,**[2] **Anu Venkatesh,**[2] **Ashwin Ram**[2]
[1]University of Maryland, Department of Computer Science and UMIACS, [2]Amazon Alexa
fenfeigo@cs.umd.edu,
{ametalli,ckhatri,ranirudh,anuvenk,ashwram}@amazon.com

## Abstract

Dialog evaluation is a challenging problem, especially for non task-oriented dialogs where conversational success is not well-defined. We propose to evaluate dialog quality using topic-based metrics that describe the ability of a conversational bot to sustain coherent and engaging conversations on a topic, and the diversity of topics that a bot can handle. To detect conversation topics per utterance, we adopt Deep Average Networks (DAN) and train a topic classifier on a variety of question and query data categorized into multiple topics. We propose a novel extension to DAN by adding a topic-word attention table that allows the system to jointly capture topic keywords in an utterance and perform topic classification. We compare our proposed topic based metrics with the ratings provided by users and show that our metrics both correlate with and complement human judgment. Our analysis is performed on tens of thousands of real human-bot dialogs from the Alexa Prize competition and highlights user expectations for conversational bots.

## 1 Introduction

Voice powered personal assistants like Amazon Alexa, Google Assistant and Apple Siri have become popular among consumers as they are intuitive to use and resemble human communication. However, most such assistants are designed for short, task-oriented dialogs, such as playing music or asking for information, as opposed to longer free-form conversations. Achieving sustained, coherent and engaging dialog is the next frontier for conversational AI, which would enable Artificially Intelligent agents to lead natural and enjoyable conversations with their human interlocutors.

Evaluation of a conversational quality for non task-oriented bots is a challenging and understudied research problem [14]. Conversational success is not well defined for such bots and, as with human-human dialogs, an interlocutor's satisfaction with a bot could be related to how engaging, coherent and enjoyable was the conversation. A conversation could be viewed as an exchange of information and opinions on a flow of topics; therefore we propose topic-based metrics to evaluate conversational bot quality. Specifically, we examine (1) Topic breadth - the ability of a bot to converse on a variety of coarse and fine grained topics without repeating itself, and (2) Topic depth - the ability to sustain long and coherent conversations on a given topic. These metrics are proposed for evaluating bots that are designed to initiate conversations on various topics and maintain user engagement in the dialog.

To classify the conversation topics, we adopt Deep Average Networks (DAN) [6] and train a supervised topic classifier on internal Question data and Alexa knowledge-query data to identify topics per utterance. Next, we extend DAN with a topic-wise attention table which learns topic-word weights across the vocabulary so that it can detect topic-specific keywords per utterance. This enables our topic classifier to not only identify the coarse level topic of an utterance such as *"Politics"*, but also to highlight fine-grained topic words like *"Trump"* which indicates the entity that the user is talking about. Both models are used to analyze tens of thousands of human-bot conversations collected as part of the Alexa Prize competition, a university competition sponsored by Amazon Alexa.

---

[*]Work done during an internship at Amazon Alexa

We show that the topic classifier and topic keyword detector achieves good accuracy on left-out internal test sets that are similar to conversational data, and can detect meaningful keywords per topic. Using the hypothesis of our topic and keyword detector we define conversational quality metrics that describe topic depth and breadth. We show that the ability of a bot to maintain long topic-specific coherent sub-conversations correlate well with user satisfaction measured by live user ratings and are therefore valid both the effectiveness of our trained topic classifier and the data-driven metrics for assessing conversation quality. Further analysis shows that some topic metrics, such as the topic-specific keywords coverage, can complement human judgment and uncover problems of a dialog system that are obscured by the user ratings due the intrinsic limitations of live data collection. Our analysis of tens of thousands of human-bot dialogs offers *automatic* and *interpretable* metrics of conversational bot quality and sheds light into *user expectations* from a good conversational bot.

## 2  Related Work

Automatic dialog evaluation is a challenging research area. Previous work has introduced evaluation metrics related to task success for evaluating task-based bots [22]. However, there is no well-defined measure of success for conversational bots, and there can be multiple valid bot responses for a given user utterance. Dialog researchers have used metrics of word overlap between a bot response and a reference valid response [19, 10, 23, 24], such as BLEU [17], METEOR [2], and embedding based metrics [24]. However, such metrics don't account well for the significant diversity between multiple valid bot responses, and have recently been shown to correlate poorly with human judgments of response quality [12]. Galley et al. [5] proposed deltaBLEU, a modified BLEU that considers multiple human-evaluated ground truth responses; however, such annotations are very expensive to collect in practice. Lowe et al. [14] proposed an Automatic Dialogue Evaluation Model (ADEM) which learns to predict human-like scores to input responses. However, ADEM requires human ratings for each response instead of conversation-level ratings, which are expensive to collect, noisy, and lack interpretability if users do not clarify their evaluation criteria. Here, we propose dialog-level *interpretable evaluation metrics* for the evaluation of topic coherence and diversity.

Deep learning advances have had a profound influence on natural language and dialog research, and researchers have experimented with Recurrent Neural Networks (RNNs)[11], [20], [13] and Convolutional Neural Networks (CNNs) [9], [4], [8] for text classification and summarization tasks. Recent work indicates that simpler bag-of-words neural models, like Deep Averaging Networks (DAN) [6] or FastText [7], can achieve state of the art results for sentence classification tasks like sentiment analysis and factoid question answering [6], and are efficient to train. In this work, we take advantage of such simple and efficient bag-of-word models for topic classification.

## 3  The Alexa Prize Competition

Alexa Prize [21] is a competition organized by Amazon Alexa machine learning [1] that aims to advance the state of conversational AI. This competition enables university research teams to innovate on AI technologies using large-scale live conversational data from real users. As part of the competition, 15 university teams were selected among hundreds of proposals and competed to win a large monetary prize to fund their research. They were provided data, funding, infrastructure, access to a large user base and expert guidance to develop *socialbots* that can converse engagingly and coherently on various popular topics and current events such as entertainment, sports, politics, and others. These 15 socialbots were deployed to the Alexa Service and released to millions of Alexa users. Alexa users could converse with a randomly selected bot on any Alexa-enabled device by using the phrase *"Alexa, let's chat"*. For each conversation, a random bot was matched with a user. At the end of the conversation, the users were prompted to rate the socialbot quality and had the option to provide feedback to the university teams.

## 4  Datasets

### 4.1  Conversational Analysis data

**Conversation data**  We analyze tens of thousands of conversations between Alexa users and the socialbots collected over one month of the competition.[2] Each conversation has a live user rating. The average number of turns per conversation is 12. Table 2 shows an example conversation.

---

[2]We used one month of data for this work, which was the data that has RER (see next page) annotations available during the timespan that this work was performed.

| Internal Question data | | Alexa knowledge-query data | |
|---|---|---|---|
| Topic | Example | Topic | Example |
| Entertainment_Movies | Who made the movie 'Interstellar'? | Politics | How old is Donald Trump? |
| Politics_Government | Can an American join the British SAS? | Sports | Who is pitching for the Red Sox? |
| Sports_Baseball | When is the Atlanta Braves next game? | Science | How far is the moon from the earth? |
| Business_Corporate | What is godaddy dot com | Phatic | How are you? |

Table 1: Left: examples from the internal Question dataset containing questions from different topics (without answers); Right: examples from topic specific knowledge queries by customers interacting with Alexa

**Live user ratings** At the end of each conversation, the user is prompted to rate the socialbot on a scale of 1 to 5 stars, based on whether they would like to chat with that socialbot again. While individual ratings are subjective, by averaging multiple ratings at the bot level we expect the average rating to indicate user satisfaction and reflect conversational bot quality. We use the live user ratings to validate our proposed topic-based metrics of conversational bot quality (Section 7.3).

**Response Error Rate (RER) for measuring coherence** To capture the coherence of the conversation, we annotated tens of thousands of randomly selected interactions for *incorrect*, *irrelevant*, or *inappropriate* responses. The Response Error Rate (RER) for each socialbot is defined as the ratio of the number of turns with erroneous responses (incorrect, irrelevant, inappropriate) over the total number of turns. Manual annotations of coherence are useful because the diversity of possible valid bot responses makes the dialog evaluation problem subjective. Here, manual RER metrics are used to *validate* the live user ratings and our proposed automatic topic-based metrics, since all of those metrics reflect dialog quality and are expected to correlate well with each other (Section 7.3).

### 4.2 Training data

The Alexa Prize conversations were not annotated in terms of topics. We used the internal Question data and Alexa knowledge-query data to train the supervised topic classifier.

**Internal Question data** This is a large Amazon internal resource containing about 5 million questions that are categorized into 55 topics. See the left side of Table 1 for some examples. This dataset provides a rich source of named entities, however it is a few years out of date.

**Alexa knowledge-query data** We annotated around 750K general knowledge queries from internal Alexa user data into 26 topics including *Science*, *Politics*, *Sports*, etc. This dataset comes from actual Alexa users and is more up-to-date. The right side of Table 1 shows a few examples. *Phatic* indicates non-topical chit-chat utterances that may include user preference ( *"I like you"*), acknowledgement (*"thanks"*, *"sounds good"*), personal questions (*"how are you"*) and other non-topical utterances.

## 5 Topic Classification

We build supervised topic classifiers to both identify the topic of each utterance and detect topic-specific keywords for the purpose of topic-based analysis of conversations. The training data, i.e. internal Question and Alexa knowledge-query data, cover a range of topics that we expect to find in bot conversations. However, the data consists mostly of questions while the human-bot utterances could also be statements, opinions, requests, greetings, etc. To prevent overfitting our classifier on the syntactic structure of the training data, we adopt the bag-of-word assumption.

### 5.1 The DAN topic classifier model

We adopt the Deep Averaging Network (DAN) model proposed by Iyyer et al. [6] as our first classifier. DAN is a bag-of-words neural model (see Figure 1) that averages the word embeddings in each input utterance as the utterance representation $s$. $s$ is passed through a series of fully connected layers, and fed into a softmax layer for classification. Assume an input utterance of length $L$, and corresponding word embeddings $e_i, i = 1, \cdots L$, then the utterance representation is: $s = \frac{1}{L}\sum_{i=1}^{L} e_i$.

DAN has been proven a state-of-the-art model for the Factoid Question Answering task [6] because of its ability to memorize keywords and its robustness to noise. This meets well our needs for topic classification, therefore DAN is an effective and efficient classifier choice. Speed is a crucial requirement, as we want to efficiently process large amounts of Alexa Prize conversations. The simple feedforward multi-layer fully-connected structure can be accelerated with the usage of GPUs.

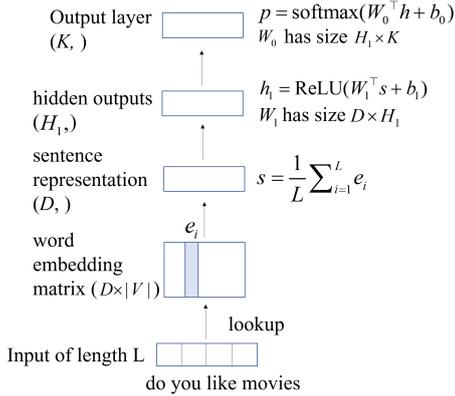
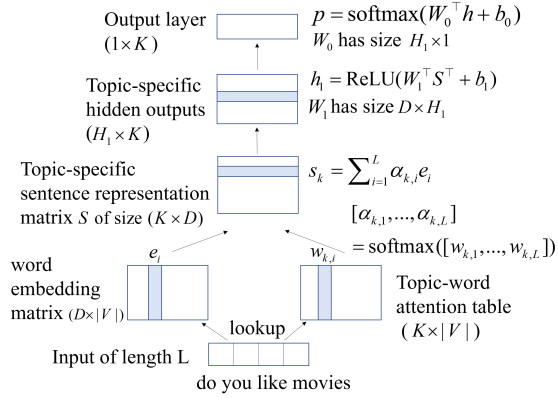

Figure 1: The structure of DAN.    Figure 2: The structure of ADAN.

### 5.2 Attentional DAN: An interpretable model for topicwise keywords detection

While the DAN model can classify the topic of an utterance, it does not explain why a certain classification decision was made. We also want to know which topic-specific keywords in each utterance relate most to its topic. For example, for *"when is the atlanta braves next game"*, we want to both classify the topic as *Sports*, and identify topic-specific keywords like *"game"* and *"braves"*. This allows a more fine-grained understanding of the topics and named entities in an utterance.

We propose an Attentional Deep Average Network (ADAN) to jointly learn the *topic-wise word saliency* for each input utterance along with the classification objective. In the probabilistic topic model, a topic is defined as a distribution over a fixed vocabulary [3]. Here, we model this distribution by a *topic-word attention table* of size $K \times |V|$ as shown in Figure 2, where $|V|$ is the vocabulary size and $K$ the number of topics in our classification task. ADAN maintains the bag-of-words assumption and applies weighted average over the input word embeddings with learned attentions to obtain the utterance representation. Specifically, the saliency weight $w_{k,i}$ in the topic-word attention table corresponds to the saliency of word $x_i$ given topic $t_k$. Assuming an utterance $[e_1, \cdots, e_L]$ of length $L$, where $e_i$ is the embedding of the $i^{th}$ word, we compute the attention weights $[\alpha_{k,1}, \cdots, \alpha_{k,L}]$ by normalizing the saliency $w_{k,i}$ with a softmax function (we normalize across utterance of length $L$ as opposed to across the whole vocabulary to reduce computation cost). The utterance representation $s_k$ per topic is computed through weighted average:

$$[\alpha_{k,1}, \cdots, \alpha_{k,L}] = \text{softmax}([w_{k,1}, \cdots, w_{k,L}]), \quad s_k = \frac{1}{L}\sum_{i=1}^{L} \alpha_{k,i} e_i \qquad (1)$$

In Figure 2, the topic-specific sentence representation $S$ is a $K \times D$ matrix where each row is a topic specific representation $s_k$ and $D$ is the embedding dimension. $S$ is passed through a series of fully connected layers and fed through a softmax. The word embedding matrix can be initialized with any pre-trained embeddings such as GloVe [18] and Word2Vec [15, 16], while the topic-word attention matrix is initialized randomly and is learnt from data. We use ReLUs as the nonlinearity. During testing, the learnt topic-word weights can be examined for each test utterance to detect the most salient words for a given topic classification decision $k$. Specifically, we examine the corresponding weight matrix $[w_{1,k}, \cdots, w_{L,k}]$ for topic $k$, rank the words according to descending weights $w_{1,k}$ and pick the top most informative words in the utterance. The ADAN model is interpretable as it allows uncovering the salient words that contribute most to the topic classification decision.

## 6 Topic Based Evaluation Metrics for Conversational Bots

We introduce interpretable dialog-level and system-level quality metrics that measure conversation topic depth and breadth. We hypothesize that if a bot is able to converse on a wide range of topics without repeating itself and can sustain lengthy conversations on a topic, it's more likely to result in an enjoyable conversation. In Section 6, we examine this hypothesis by computing correlations of our metrics with the live user ratings collected in the Alexa Prize competition. Below, we define the following concepts as the basis of proposed topic related metrics.

**Topic-specific turn $\mathcal{T}$:** Defined as a pair of user utterance and conversational bot response where both utterances belong to the same topic, excluding *Phatic*. For example turns 1, 2, 3 in Table 2 are all topic-specific turns about *Music*. *Phatic* is excluded as it does not have a defined topic, and can be assumed to be chit-chat in the context of the current topic.

**Topic coherent sub-conversation** $\mathcal{S}$: Defined as a series of consecutive topic-specific turns which contain at least two consecutive turns, and the *Phatic* turn is excluded. For example, turns 1-3 in Table 2 is a sub-conversation on *Music*.

**Length of sub-conversation** $l_s$: Defined as the number of topic-specific turns it contains. For example, the *Music* sub-conversation of Table 2 has length 3.

**Length of consecutive conversation** $l_c$: Defined as the total number of the topic-specific turns. For example, the conversation of Table 2 has length 4.

Table 2: Example socialbot conversation turns and corresponding topics

| Turns | Utterances | Topics |
| --- | --- | --- |
| 1. User | Let's talk about music | Music |
| 1. Socialbot | Sure, what's your favorite musician? | Music |
| 2. User | Bob Dylan | Music |
| 2. Socialbot | Bob Dylan is an American songwriter, singer, painter, and writer. | Music |
| 3. User | Cool | Phatic |
| 3. Socialbot | Do you want to know more about Bob Dylan? | Music |
| 4. User | No, let's talk about politics instead | Politics |
| 4. Socialbot | Sure, here are the latest updates about Donald Trump [...] | Politics |

## 6.1 Conversational topic depth

By the definition of a sub-conversation, the more consecutive turns a sub-conversation has on a specific topic, the deeper the conversation goes into this topic. Suppose a conversation $\mathcal{C}_i$ has $m$ sub-conversations $\mathcal{S}_{ij}, j = 1, ..., m$, we can define:

**Dialog-level average topic depth** $D(\mathcal{C}_i)$ as the average length of all sub-conversations in each user-bot conversation: $D(\mathcal{C}_i) = \frac{1}{m} \sum_{j=1}^{m} l_s(\mathcal{S}_{ij})$

**System-level average topic depth** $D(\mathcal{B})$ is the average length of all sub-conversations $\mathcal{C}_i, i = 1, ..., n$ generated by a conversational bot $\mathcal{B}$.

## 6.2 Conversational topic breadth

Conversational topic breadth measures the diversity of topics in each dialog. We measure diversity of both coarse-grain topics like *Sports* and fine-grain topic words like *"yankees"*.

### 6.2.1 Coarse topic domain coverage

**Dialog-level coarse topic breadth** $Br(\mathcal{C}_i)$ is the number of distinct topics $t_k$ that occur during at least one sub-conversation in conversation $\mathcal{C}_i$. A large $Br(\mathcal{C}_i)$ means the user chats with the bot about multiple topics and each topic is sustained for at least 2 turns, which may indicate an enjoyable conversation.

**System-level coarse topic breadth** $Br_{avg}(\mathcal{B}) = \frac{1}{n} \sum_{i=1}^{n} Br(\mathcal{C}_i)$, where $n$ is the number of conversations generated by bot $\mathcal{B}$, shows the general ability of $\mathcal{B}$ to engage in diverse conversations.

**System-level coarse topic count** $N(\mathcal{B}, t_k)$ is the total number of sub-conversations with topic $t_k$ in all conversations $\mathcal{C}_i$ by $\mathcal{B}$. The histogram of $N(\mathcal{B}, t_k)$ across topics shows whether a bot mostly talks about a few topics or tends to talk equally about a range of topics. By normalizing this histogram, we obtain the **System-level coarse topic frequency** $F(\mathcal{B}, t_k)$. A low standard deviation or high entropy of $F(\mathcal{B}, t_k)$ shows that $\mathcal{B}$ has a balanced ability to talk across a range of topics.

### 6.2.2 Topic-specific keywords coverage

The system-level topic-specific keyword coverage metrics are defined to measure the ability of a conversational bot to cover a variety of fine-grained topics without repeating itself across dialogs. For example, a bot that always talks about *"Dunkirk"* as part of a *Movies* conversation would probably not be enjoyable to users that interact regularly with it. We use the top-2 most salient keywords per utterance detected by ADAN (Section 5.2) to compute the following metrics.

**System-level topic keyword coverage** $C_{fine}(\mathcal{B})$ is the total number of distinct topic keywords across all conversations $\mathcal{C}_i$ generated by conversational bot $\mathcal{B}$.

**System-level topic keyword count** $N(\mathcal{B}, w_k)$ is the total count of topic keywords $w_k$ across all conversations $\mathcal{C}_i$ by $\mathcal{B}$. By averaging over topic keywords, we get the **System-level topic keyword frequency** $F(\mathcal{B}, w_k)$.

# 7 Results and Discussion

We evaluate our proposed topic classifiers by separating both internal Question data and Alexa knowledge-query data (see Section 4.2) into training, dev and test sets by the ratio 8:1:1.

## 7.1 Topic classification with DAN

By varying the network structure we find no significant improvements with a network of more than 1 hidden layer. Therefore, to prevent overfitting and ensure data processing efficiency, we fix our network structure with an embedding dimension of 300 and one hidden layer with 500 nodes.

The results on the internal Question test data are presented in Table 3, where the first column shows classification accuracy of the 55 fine grained topics, while the second column shows accuracy when merging similar topics into 26 coarser topics (e.g., *Sports_Baseball*, *Sports_Football* and *Sports_Scores* were merged into *Sports*). Accuracy reaches over 80% for the latter case.

| Method | Accuracy | |
| --- | --- | --- |
| | 55 topics | 26 merged topics |
| Random initialization | 71.2% | – |
| Fixed GloVe [18] initialization | 75.6% | 77.1% |
| GloVe [18] + fine tune | 77.3% | **82.4%** |

Table 3: DAN model topic classification accuracy on the internal Question test data

We also evaluate DAN on the Alexa knowledge-query data which are annotated into 26 topics. As described in Section 4, this data set contains 750K annotated utterances, 10% of which are left as a test set, and 10% as a dev set. The data is highly imbalanced in topics, e.g., 32.7% of utterances are *Phatic* whereas only 1.2% are annotated as *Literature*. In Table 4 we show test set accuracy Overall, for Phatic and Non-phatic classes separately, and for Literature as an example under-represented topic. To achieve more balanced accuracy across topics we tried downsampling the Phatic class and transfer learning, e.g., transferring the embeddings and weights of hidden layers from a network trained on the larger internal Question dataset. For the latter, the final output layer was replaced to be consistent with the Alexa knowledge-query labels, and the network was fine tuned on the Alexa knowledge-query data. That way, we are able to use both the internal Question and Alexa knowledge-query data, and achieve the highest accuracy overall and reasonable accuracy for under-represented topics, see Table 4. We applied this last model for Alexa Prize topic classification.

## 7.2 Topic-specific keywords identification with ADAN

In section 5.2, we propose the ADAN model that jointly classifies topics and detects salient topic-specific keywords for each utterance. We evaluate ADAN on the internal Question data, and observe a slight accuracy decrease compared to DAN (82.4% vs 80.6% on 26 Question topics). This can be attributed to the large increase of randomly initialized parameters, e.g., attention table, for ADAN. Detailed results are omitted for lack of space. We tolerate this decrease as ADAN offers interpretability and keyword detection. We examine the keyword detection quality by looking at the top-2 keywords for the most likely topic for various input utterances, as in Table 5, and observe reasonable model behavior. We have manually verified the detected keyword quality for many example utterances, while in future we plan to perform more systematic evaluation. We also examine representative topic keywords overall, by computing top-2 keywords per utterance for the internal Question test data, and visualizing the most frequent keywords per topic, in Table 6. Again, we observe that the most frequent keywords relate highly to their topics.

| Method | Accuracy | | | |
| --- | --- | --- | --- | --- |
| | Overall 26 category | Phatic | Non-Phatic | Literature |
| Randomly initalized word embeddings | 68.2% | 87.5% | 58.9% | 27.2% |
| GloVe +fine tune | 75.6% | 88.3% | 61.8% | 34.6% |
| GloVe +fine tune + downsample Phatic | 71.3% | 79.1% | **67.5%** | **43.3%** |
| Transfer weights from Question data + fine tune | **77.3%** | 88.2% | 64.8% | 41.9% |

Table 4: DAN model topic classification accuracy on Alexa knowledge-query test data

| Example Utterances | Topic | Top-2 keywords |
|---|---|---|
| "My favorite show is Star Trek" | Movies_TV | show, Trek |
| "What are the ingredients for apple pie" | Food_Drink | apple, ingredients |

Table 5: Examples of topic classification and corresponding topic keywords by ADAN for input utterances

| Topic | Most frequent keywords |
|---|---|
| Shopping | cost, walmart, price, worth, gift, stores, cheapest,... |
| Religion | bible, life, believe, love, catholic, christian, jewish,... |

Table 6: Examples of most frequent topic words for the internal Question test data obtained by ADAN

### 7.3 Conversational analysis results

We apply DAN to tens of thousands of human-bot conversations across all 15 socialbots from the Alexa Prize competition and obtain the hypothesis topics for both user utterances and socialbot responses, and use ADAN to obtain the topic-specific keywords. We use the *ensemble results* of the two best performing classifiers trained on the internal Question data and Alexa knowledge-query data respectively. Specifically, for a given utterance, we compare the *normalized entropy* of the two classification results and use the one with lower entropy, which indicates higher confidence. While there is no ground truth for the conversation data, through manual observation we noticed that ensemble results are more robust than that of a single classifier. Ensembling allows us to combine the merits of both datasets: the internal Question data which contains richer named entities, and the Alexa knowledge-query data which are more current. Ensembling also provides richer topics, e.g., the internal Question data doesn't contain *Phatic* while the Alexa knowledge-query data doesn't contain *Games*, and by ensembling we can detect both.

We noticed that the live user ratings collected for each conversation are very noisy and have high average standard deviation within each bot (~1.52). Therefore, we decided to compare our evaluation with human judgment at the system level. To have a reference for comparison, we compute the Spearman's rank correlation coefficient $\rho$ between the *aggregated live user ratings* and the *Response Error Rate (RER)*. From Table 7, we can see that the RER correlates well negatively with the user ratings for 15 chatbots ($\rho = -0.717$). This is reasonable since the fewer erroneous responses there are, a higher user rating is expected. Therefore, we can use the absolute value $|\rho| = 0.717$ as a reference to evaluate our proposed topic metrics. The analysis below presents correlations of individual metrics with user ratings, while combination of multiple metrics is left as future work.

#### 7.3.1 Evaluating conversational topic depth

We assume that the longer the conversation goes on a specific topic, the deeper that topic is explored and the more satisfied the user is. To examine this assumption, we compare the system-level topic depth $D(\mathcal{B})$, i.e., the average length of topic-specific sub-conversations, with the human judgments in Table 7. We can see that its correlation with the user ratings is almost as high as the absolute correlation of RER ($|\rho_{RER}| = 0.707$) with human. To highlight the importance of the topic-specific metric, we compute the correlation for the average length of the whole conversations (regardless of topics) as well. From Table 7, we can see that the topic depth correlates better with the human judgment than the conversation length. This demonstrates our assumption, i.e., although generally longer conversations are preferred ($\rho = 0.575$), the user is more satisfied if the bot is able to sustain longer sub-conversations on the same topic. We can use $D(\mathcal{B})$ as a proxy for user satisfaction. These results also show the effectiveness of our trained topic classifier for Alexa Prize conversations.

| Metrics | Annotated RER | Avg. sub-conversation length $D(\mathcal{B})$ | Avg. conversation length |
|---|---|---|---|
| $\rho$ | -0.717 | 0.707 | 0.575 |

Table 7: Correlation between topic depth metrics and human judgments across 15 socialbots within one month. Each Spearman's correlation score $\rho$ is computed with system-level aggregated live user ratings

#### 7.3.2 Evaluating conversational topic breadth

**A. Coarse-grain topic breadth evaluation**

We compare the topic breadth metrics of Section 6.2 with the human judgments in Table 8. The average topic breadth $Br_{avg}(\mathcal{B})$ within each conversation correlates reasonably well with user ratings ($\rho = 0.512$), which means that when users chat with a bot about various topics in the same conversation, they tend to rate the bot higher. However coarse topic frequency $F(\mathcal{B}, t_k)$, which describes the bot's ability to chat about topics in a balanced way as opposed to preferring certain topics, does not correlate highly with user rating ($\rho = 0.291$). This could be a side-effect of the competition protocol, where users are randomly matched with one of 15 bots for each conversation, so most users would only rate a specific bot a handful of times at most and may not notice if a bot is repetitive across conversations. However, avoiding repetition is important when a conversational bot interacts frequently with the same user, which is a common commercial use case. Our breadth metric is capturing complementary information to the user ratings, as it is explicitly designed to describe repetitiveness.

| Metrics | Annotated RER | Avg. coarse topic breadth $Br_{avg}(\mathcal{B})$ | Entropy of coarse topic frequency $F(\mathcal{B}, t_k)$ |
|---|---|---|---|
| $\rho$ | -0.717 | 0.512 | 0.291 |

Table 8: Correlation between topic breadth metrics and human judgments across 15 socialbots within one month. Each Spearman's correlation score $\rho$ is computed with system-level aggregated live user ratings

### B. Fine-grain topic breadth evaluation

To detect if a bot repeats itself about topic-specific named entities, we design fine-grained topic breadth metrics (see Section 6.2). Table 9 compares the behavior of bots and users in the human-bot dialogs for the top-6 ranked socialbots in the competition using the fine-grain metrics $C_{fine}(\mathcal{B})$ and $F(\mathcal{B}, w_k)$. Although the bot ratings are reasonably good (over 3 out of 5 stars), their behaviors are quite different. For example, $\mathcal{B}_6$ has a lower topic keywords coverage $C_{fine}(\mathcal{B})$ and a higher topic keyword frequency $F(\mathcal{B}, w_k)$, which indicates that this bot tends to talk about a narrower range of fine-grained topics and repeats itself more often. Note that the corresponding metrics of the user behavior are relatively stable across the bots, e.g., users do not constrain their conversation to certain fine-grained topics and respond with a broad vocabulary in general. Similarly to the coarse frequency $F(\mathcal{B}, t_k)$, the fine topic word frequency $F(\mathcal{B}, w_k)$ describes repetitiveness at a fine-grained keyword level. As described earlier, repetitiveness is not well captured by the user ratings in our setup, since a typical user gets to rate just a few random bots, and may notice if a bot repeats itself.

|  |  | $\mathcal{B}_1$ | $\mathcal{B}_2$ | $\mathcal{B}_3$ | $\mathcal{B}_4$ | $\mathcal{B}_5$ | $\mathcal{B}_6$ |
|---|---|---|---|---|---|---|---|
| Average live user ratings | – | 3.51 | 3.38 | 3.19 | 3.06 | 3.21 | 3.13 |
| Topic keyword frequency $F(\mathcal{B}, w_k)$ | Bot | 9.49 | 6.23 | 10.21 | 8.86 | 12.35 | **20.56** |
| Topic keyword frequency $F(\mathcal{B}, w_k)$ | User | 5.98 | 7.00 | 6.69 | 5.29 | 5.72 | 6.09 |
| Topic keyword coverage $C_{fine}(\mathcal{B})$ | Bot | 2469 | 3159 | 2562 | 3022 | 1944 | **904** |
| Topic keyword coverage $C_{fine}(\mathcal{B})$ | User | 3041 | 1859 | 2976 | 2520 | 1576 | 2247 |

Table 9: Comparing the bot and user behavior for top 6 ranked socialbots by fine-grained topic breadth metrics. In the second column,"Bot" indicates that the metrics are computed with purely bot responses in all dialogs; "User" indicates that only user utterances are used.

## 8 Conclusions and Future Work

The purpose of this work is to design interpretable metrics for evaluating non-task chat-oriented dialog systems and provide meaningful guidance for designing the pipeline of a conversational AI. Along these lines, we proposed several dialog-level/system-level evaluation metrics based on topic coherence, diversity and depth. These metrics are designed for evaluating conversational bots that are developed to lead and sustain engaging conversations on a variety of topics. We developed a novel attention-based neural topic model, the Attentional Deep Average Network (ADAN), that jointly learns a topic-word attention table along with the classification objective. This enabled us to identify both the most likely topic and the most salient topic-specific keywords in each utterance. Our model was trained and evaluated on a variety of internal datasets. Our results demonstrate good topic classification accuracy and the ADAN model's ability to identify reasonable topic-specific keywords.

Using the results of both DAN [6] and ADAN and our proposed topic-based metrics, we evaluated multiple aspects of a bot's quality including for how long it can sustain coherent sub-conversation on a topic and whether it can lead non-repetitive dialogs on a variety of fine and coarse grained topics. We showed that a user's satisfaction correlates well with long and coherent on-topic conversations, while metrics of topic breadth may provide complementary information to user ratings, as the repetitiveness of topics is hardly captured in user ratings due to the intrinsic limitations of live user data collection. Future work will address how to incorporate such metrics for improving the quality of a conversational bot. We also plan to investigate unsupervised topic modeling methods to learn latent topics from large amounts of unlabeled human-bot interactions as well as develop topic models that incorporate conversation context from both interaction parties.

# References

[1] Amazon Alexa Machine Learning. https://www.amazon.jobs/en/teams/alexa-machine-learning.

[2] S. Banerjee and A. Lavie. METEOR: An automatic metric for MT evaluation with improved correlation with human judgments. In *Proceedings of the ACL workshop on intrinsic and extrinsic evaluation measures for machine translation and/or summarization*, 2005.

[3] D. M. Blei. Probabilistic topic models. *Communications of the ACM*, 55(4):77–84, 2012.

[4] P. Blunsom, E. Grefenstette, N. Kalchbrenner, et al. A convolutional neural network for modelling sentences. In *Proceedings of the 52nd Annual Meeting of the Association for Computational Linguistics*, 2014.

[5] M. Galley, C. Brockett, A. Sordoni, Y. Ji, M. Auli, C. Quirk, M. Mitchell, J. Gao, and B. Dolan. deltaBLEU: A discriminative metric for generation tasks with intrinsically diverse targets. In *Proceedings of the Annual Meeting of the Association for Computational Linguistics and the International Joint Conference on Natural Language Processing*, 2015.

[6] M. Iyyer, V. Manjunatha, J. Boyd-Graber, and H. Daumé III. Deep unordered composition rivals syntactic methods for text classification. *Proceedings of the 53rd Annual Meeting of the Association for Computational Linguistics*, 2015.

[7] A. Jiulin, E. Grave, P. Bojanowski, and T. Mikolov. Bag of tricks for efficient text classification. In *Proceedings of the 15th Conference of the European Chapter of the Association for Computational Linguistics)*, page 427–431, 2017.

[8] Y. Kim. Convolutional neural networks for sentence classification. *Proceedings of EMNLP*, 2014.

[9] S. Lai, L. Xu, K. Liu, and J. Zhao. Recurrent convolutional neural networks for text classification. In *Twenty-Ninth AAAI Conference on Artificial Intelligence*, 2015.

[10] J. Li, M. Galley, C. Brockett, G. P. Spithourakis, J. Gao, and B. Dolan. A persona-based neural conversation model. In *Proceedings of the 54th Annual Meeting of the Association for Computational Linguistics*, page 994–1003, 2016.

[11] R. Lin, S. Liu, M. Yang, M. Li, M. Zhou, and S. Li. Hierarchical recurrent neural network for document modeling. In *Proceedings of EMNLP*, pages 899–907, 2015.

[12] C.-W. Liu, R. Lowe, I. V. Serban, M. Noseworthy, L. Charlin, and J. Pineau. How not to evaluate your dialogue system: An empirical study of unsupervised evaluation metrics for dialogue response generation. *arXiv preprint arXiv:1603.08023*, 2016.

[13] P. Liu, X. Qiu, X. Chen, S. Wu, and X. Huang. Multi-timescale long short-term memory neural network for modelling sentences and documents. In *Proceedings of the 2015 Conference on Empirical Methods in Natural Language Processing*, 2015.

[14] R. Lowe, M. Noseworthy, I. V. Serban, N. Angelard-Gontier, Y. Bengio, and J. Pineau. Towards an automatic turing test: Learning to evaluate dialogue responses. Proceedings of ACL, 2017.

[15] T. Mikolov, K. Chen, G. Corrado, and J. Dean. Efficient estimation of word representations in vector space. *arXiv preprint arXiv:1301.3781*, 2013.

[16] T. Mikolov, I. Sutskever, K. Chen, G. S. Corrado, and J. Dean. Distributed representations of words and phrases and their compositionality. In *Proceedings of NIPS*, 2013.

[17] K. Papineni, S. Roukos, T. Ward, and W.-J. Zhu. BLEU: a method for automatic evaluation of machine translation. Proceedings of ACL, 2002.

[18] J. Pennington, R. Socher, and C. D. Manning. Glove: Global vectors for word representation. In *Proceedings of EMNLP*, 2014.

[19] A. Sordoni, M. Galley, M. Auli, C. Brockett, Y. Ji, M. Mitchell, J.-Y. Nie, J. Gao, and B. Dolan. A neural network approach to context-sensitive generation of conversational responses. In *Proceedings of the Conference of the North American Chapter of the Association for Computational Linguistics (NAACL-HLT 2015).*, 2015.

[20] D. Tang, B. Qin, and T. Liu. Document modeling with gated recurrent neural network for sentiment classification. In *Proceedings of the 2015 Conference on Empirical Methods in Natural Language Processing*, pages 1422–1432, 2015.

[21] The Alexa Prize Competition. https://developer.amazon.com/alexaprize.

[22] M. Walker, C. Kamm, and D. Litman. Towards developing general models of usability with PARADISE. *Natural Language Engineering*, 6(3-4):363–377, 2000.

[23] T.-H. Wen, M. Gasic, N. Mrksic, P.-H. Su, D. Vandyke, and S. Young. Semantically conditioned LSTM-based natural language generation for spoken dialogue systems. In *Proceedings of the 2015 Conference on Empirical Methods in Natural Language Processing*, page 1711–1721, 2015.

[24] T. Zhao, R. Zhao, and M. Eskenazi. Learning discourse-level diversity for neural dialog models using conditional variational autoencoders. In *Proceedings of the 55th Annual Meeting of the Association for Computational Linguistics*, page 654–664, 2017.